\documentclass[a4paper,conference]{IEEEtran}
\usepackage[utf8]{inputenc}
\usepackage[T1]{fontenc}
\usepackage{cite}
\usepackage{graphicx}
\usepackage{amsmath,amssymb}
\usepackage{booktabs}
\usepackage{hyperref}
\usepackage{multirow}
\usepackage{array}
\usepackage{caption}
\usepackage[dvipsnames]{xcolor}
\usepackage{tikz}
\usepackage{pgf,pgfplots}
\usetikzlibrary{positioning}
\usepackage{subcaption}
\usepackage{lipsum}
\usepackage{algorithm}
\usepackage{algpseudocode}
\usepackage{siunitx}
\usepackage{makecell}
\usepackage{adjustbox}
\usepackage{cleveref}
\usepackage{float}
\usepackage{balance}
\usepackage{todonotes}
\usepackage{adjustbox}

\usepackage{pgfplotstable}
\pgfplotsset{compat=1.18}
\begin{document}

\title{FedSparQ: Adaptive Sparse Quantization with Error Feedback for Robust \& Efficient Federated Learning}

\author{\IEEEauthorblockN{Chaimaa MEDJADJI}
\IEEEauthorblockA{\textit{University of Luxembourg, Luxembourg} \\
chaimaa.medjadji@uni.lu}
\and
\IEEEauthorblockN{Sadi ALAWADI}
\IEEEauthorblockA{\textit{Blikinge Institute of Technology, Sweden} \\
sadi.alawadi@bth.se}
\and
\IEEEauthorblockN{Feras M. Awaysheh}
\IEEEauthorblockA{\textit{ADSLab, Umea University, Sweden} \\
feras.awaysheh@umu.se}
\and
\IEEEauthorblockN{Guilain LEDUC}
\IEEEauthorblockA{\textit{University of Luxembourg, Luxembourg} \\
guilain.leduc@uni.lu}
\and
\IEEEauthorblockN{Sylvain KUBLER}
\IEEEauthorblockA{\textit{University of Luxembourg, Luxembourg} \\
sylvain.kubler@uni.lu}
\and
\IEEEauthorblockN{Yves Le Traon}
\IEEEauthorblockA{\textit{University of Luxembourg, Luxembourg} \\
Yves.LeTraon@uni.lu}
}
\maketitle

\begin{abstract}
Federated Learning (FL) enables collaborative model training across decentralized clients while preserving data privacy by keeping raw data local. However, FL suffers from significant communication overhead due to the frequent exchange of high-dimensional model updates over constrained networks.
In this paper, we present \textbf{FedSparQ}, a lightweight compression framework that dynamically sparsifies the gradient of each client through an adaptive threshold, applies half-precision quantization to retained entries and integrates residuals from error feedback to prevent loss of information. FedSparQ requires no manual tuning of sparsity rates or quantization schedules, adapts seamlessly to both homogeneous and heterogeneous data distributions, and is agnostic to model architecture. Through extensive empirical evaluation on vision benchmarks under independent and identically distributed (IID) and non-IID data, we show that FedSparQ substantially reduces communication overhead (reducing by 90\% of bytes sent compared to FedAvg) while preserving or improving model accuracy (improving by 6\% compared to FedAvg non-compressed solution or to state-of-the-art compression models) and enhancing convergence robustness (by 50\%, compared to the other baselines). Our approach provides a practical, easy-to-deploy solution for bandwidth-constrained federated deployments and lays the groundwork for future extensions in adaptive precision and privacy-preserving protocols.
\end{abstract}


\section{Introduction}
Federated Learning (FL)~\cite{mcmahan2017Communication} enables distributed edge nodes (clients and workers) to collaboratively train machine learning models locally, preserving data privacy by sharing only model updates instead of raw data. However, in practical deployments (e.g.\ mobile phones, IoT sensors, or geographically dispersed edge servers, ...), communication bandwidth is severely constrained and often the dominant bottleneck \cite{awayshehBig}. Transmitting full-precision gradients or parameters each round can exhaust network quotas, increase latency, and even drain device batteries. Moreover, client‐side heterogeneity (in compute power, data distribution, and network connectivity) exacerbates these challenges, since a ``one‐size‐fits‐all" compression strategy can either underutilize bandwidth when updates are small or discard important information when updates are large.  



\begin{figure}[t!]
\centering
\includegraphics[width=0.9\linewidth]{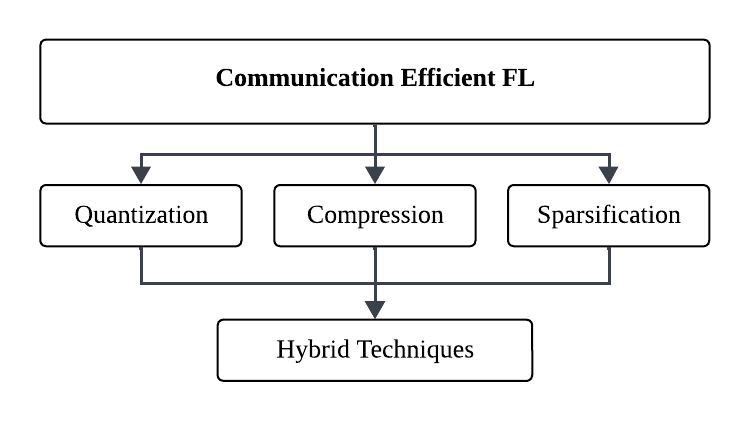}
\caption{Taxonomy of communication-efficient federated learning (FL). 
The top node states the goal; the middle layer distinguishes three complementary families:
(i) \emph{Quantization}; (ii) \emph{Sparsification};
(iii) \emph{Compression (scheduling/physical-layer)}. 
Arrows to \emph{Hybrid Techniques} indicate methods that combine multiple techniques to obtain multiplicative savings while preserving convergence.}
\label{fig:CEFLTech}
\end{figure}

Several approaches exist to mitigate this communication overhead problem. \figurename~\ref{fig:CEFLTech} summarizes the three main approaches:

To overcome this limitation, we propose \textbf{FedSparQ}, a unified compression framework that: (1) performs \emph{FedProx-regularized local updates} to avoid client's model drift on non-IID data, (2) uses \emph{EMA-smoothed, layer-wise thresholds} to select exactly the most informative parameters each round, (3) quantizes only those retained updates to float16, halving the per-value byte cost, and  (4) uses error-feedback with residual accumulation to compensate for the bias introduced by dropped updates. Our Key contributions are:

\begin{itemize}
    \item \textbf{Communication-efficient:} using an adaptive compression strategy that combines exponentially smoothed, layer-wise sparsification with float16 quantization. This compound approach selects only the most informative gradient coordinates and encodes them with reduced precision, achieving significant bandwidth savings without manual tuning.
    
    \item \textbf{Robustness to non-IID data:} by integrating FedProx regularization to mitigate client drift and stabilize optimization in the presence of skewed and heterogeneous data distributions, a common challenge in real-world FL settings.
    
    \item \textbf{Unbiased convergence via residual feedback:} by employing an error-feedback mechanism to accumulate and re-inject discarded gradient information, ensuring convergence remains unbiased despite aggressive gradient pruning.
\end{itemize}

Several experiments have been conducted across various datasets, including MNIST, FashionMNIST, and CIFAR-10. Our results show that FedSparQ achieves up to a 90\% reduction in communication overhead without hurting and, in some cases, improving model accuracy across varied non-IID settings and heterogeneous devices.

The remainder of this paper is organised as follows:  Section~\ref{Sec:Backgound} provides a general background about FedAvg, Quantization, Sparsification and error-feedback residual. Section~\ref{Sec:rw} provides a comprehensive overview of previous research addressing FL communication efficiency challenges. The proposed FedSparQ approach is introduced and detailed in Section~\ref{Sec:fedsparq}, including its functional mechanism and theoretical motivation. Section~\ref{Sec:conv} presents the formal convergence analysis of the proposed approach. Experimental evaluations and found results across three benchmark datasets and six baseline methods are reported in Section~\ref{Sec:Results} followed by a discussion of the implications and insights derived from the results, and the main conclusions and future directions are summarized in Section~\ref{Sec:conclu}.

\section{Background}
\label{Sec:Backgound}

This section introduces FL as a privacy-by-design \cite{ZeroTrustFL} and FedAvg, the foundational FL algorithm that works without compression techniques. In addition, we briefly explain key compression techniques such as quantisation, sparsification, and error-feedback, which aim to reduce communication overhead in FL systems. These techniques form the conceptual basis for the proposed FedSparQ approach.
\subsection{FedAvg}\label{Sec:Background:FedAvg}

Consider a federation of \(K\) clients, where client \(k\) holds a local dataset \(\mathcal{D}_k\) of size \(n_k\), and the total number of examples is \(N=\sum_{k=1}^K n_k\). Federated learning seeks to minimize the global empirical risk:
\[
  F(w) \;=\; \frac{1}{N}\sum_{k=1}^K \sum_{(x,y)\in\mathcal{D}_k} \ell(w; (x,y))
  \;=\;\sum_{k=1}^K \frac{n_k}{N}\,F_k(w)\,,
\]
where \(F_k(w)\) is client \(k\)'s local loss.  

The canonical \emph{FedAvg}~\cite{mcmahan2017Communication}  algorithm proceeds in communication rounds \(t=1,2,\dots\). At round \(t\):
\begin{enumerate}
  \item The server broadcasts the current global model \(w_t\) to a subset \(S_t\) of clients.
  \item Each client \(k\in S_t\) performs one or more steps of SGD on its local objective \(F_k\), yielding an updated model \(w_t^k\).
  \item Clients return their updates \(\Delta_t^k = w_t - w_t^k\) to the server.
  \item The server aggregates:
  \[
    w_{t+1} = w_t - \eta \sum_{k\in S_t} \frac{n_k}{\sum_{j\in S_t} n_j}\,\Delta_t^k,
  \]
  where \(\eta\) is the global learning rate.
\end{enumerate}

While FedAvg preserves data privacy, each round incurs a communication cost of \(O(d)\) parameters (32-bits each) per participating client, which can overwhelm bandwidth‐constrained devices.

\subsection{Quantization}
\label{Sec:Background:Quant}
Quantization compresses each gradient or parameter update by reducing its bit‐width. A common unbiased quantizer \(Q_s:\mathbb{R}^d\to\mathbb{R}^d\) (e.g.\ QSGD~\cite{alistarh2017qsgd}) operates coordinate‐wise:
\[
  [Q_s(x)]_i \;=\; \|x\|_2\,\mathrm{sign}(x_i)\,\xi_i,
  \quad
  \mathbb{E}[\xi_i] = \frac{|x_i|}{\|x\|_2},
\]
ensuring \(\mathbb{E}[Q_s(x)] = x\).  Mixed‐precision quantization simply casts float32 values to float16:
\[
  Q_{16}(x_i) = \mathrm{float16}(x_i),
\]
halving per‐coordinate bytes at the cost of added rounding noise.

\subsection{Sparsification}
\label{Sec:Background:Sparse}
Sparsification sends only a subset of coordinates.  Two primary schemes are:
\begin{itemize}
  \item \emph{Static top-\(k\)}: retain the \(k\) largest‐magnitude entries:
  \[
    [S_k(x)]_i = 
    \begin{cases}
      x_i, & i\in \mathrm{TopK}(|x|),\\
      0,   & \text{otherwise},
    \end{cases}
  \]
  guaranteeing exactly \(k\) values per update.

  \item \emph{Threshold sparsification}~\cite{lin2018deep}: choose a threshold \(\tau\), often \(\tau = \frac{1}{d}\sum_{i}|x_i|\), and keep
  \[
    [S_\tau(x)]_i = x_i \,\mathbf{1}\{|x_i|\ge\tau\}.
  \]
  This \emph{dynamic} sparsifier adapts the support size \(|I(x)|\) to the gradient distribution each round.
\end{itemize}

Sparsification reduces communication to \(O(|I|)\) non-zero entries, but discarding small entries introduces a bias that can slow or even prevent convergence.

\subsection{Error‐Feedback Residuals}
\label{Sec:Background:EFR}
Error‐feedback (EF) addresses the bias of compression by accumulating the dropped ``residual'':
\[
  r_{t+1} = x_t - C(x_t)\,,\quad
  \tilde x_t = C(x_t + r_t),
\]
where \(C\) is a compressor (quantizer or sparsifier).  In the next round, \(r_t\) is added back into the input, ensuring that \(\sum_t \tilde x_t = \sum_t x_t\) in expectation, and restoring the convergence properties of uncompressed SGD~\cite{lin2018deep}.

\subsection{Hybrid Compression Methods}



Hybrid FL methods combine sparsification and quantization, often with periodic averaging or control variates.

However, these methods typically rely on multiple hyperparameters and do not adapt their compression levels to per‐round gradient statistics.

FedSparQ builds on these foundations by unifying \emph{adaptive threshold} sparsification, float16 quantization, and error-feedback residual accumulation into a single, hyperparameter-light framework that automatically balances communication reduction against unbiased convergence.


\section{Related Work}
\label{Sec:rw}
To mitigate the communication bottleneck in FL, researchers have developed a spectrum of compression techniques that combine both quantization and sparsification methods, augmenting them sometimes with Error-Feedback residual accumulation to correct bias and preserve convergence.

\textbf{Quantization Techniques.} Gradient and parameter quantization shrink message sizes by reducing numeric precision. For example, QSGD \cite{alistarh2017qsgd} offers an unbiased stochastic quantizer with provable guarantees, while Deep Gradient Compression \cite{lin2018deep} unites thresholding, momentum correction, and quantization to slash communication costs in centralized settings. Adapting these techniques to FL requires accounting for client heterogeneity and privacy constraints. Simple one-bit or low-bit schemes remain popular for their ease and efficacy. The SOBAA framework \cite{oh2024sobaa} merges sparse one-bit quantization with over-the-air analog aggregation and power control to balance convergence speed against bandwidth. Building on this, FedFQ \cite{li2024fedfq} employs a simulated-annealing approach to assign bit-widths at the parameter level, further refining the trade-off between precision and communication. Complementary to these, \emph{SoteriaFL} couples stochastic compression with local differential privacy and analyzes privacy–utility trade-offs for compressed FL \cite{li2022soteriafl}.

\textbf{Sparse Communication.} Sparsification techniques prune less significant updates to shrink communication payloads. Early methods like random sparsification~\cite{aji2017sparse, wangni2018gradient}, which transmits a fixed random subset of gradients, and static top-$k$ sparsification~\cite{dryden2016communication, lin2017deep}, which selects the largest magnitudes, require manual selection of sparsity rates and do not adapt to evolving training dynamics. Hybrid federated frameworks such as FedPAQ~\cite{reisizadeh2020fedpaq} and Sparse Ternary Compression~\cite{sattler2019sparse} combine quantization, sparsification, and periodic averaging to achieve high compression, yet still depend on fixed hyperparameters. More recent work addresses this rigidity: SpaFL~\cite{kim2024spafl} learns structured, trainable thresholds to prune entire filters and neurons, communicating only the thresholds themselves. By dynamically adjusting sparsity patterns according to the model’s state, SpaFL significantly reduces both communication and computation costs while preserving accuracy, an adaptive principle we extend in FedSparQ. In the federated setting, \emph{SparseFed} further explores sparsity-driven aggregation with robustness and convergence analysis tailored to device heterogeneity \cite{panda2022sparsefed}.

\textbf{Residual Feedback.} To counteract the accuracy degradation caused by aggressive compression, Residual Feedback~\cite{karimireddy2019error} mechanisms accumulate and retransmit quantization errors in subsequent rounds. This technique enhances convergence and stability, especially when integrated with sparsification and quantization~\cite{lin2018deep}.

\textbf{Hybrid techniques.} While earlier works treated sparsification, quantization, and residual‐feedback separately, more recent hybrid methods jointly optimize these components. For example, JointSQ~\cite{li2024jointsq} interprets sparsification as 0‐bit quantization and employs mixed‐bit precision within a unified framework, and FedPAQ~\cite{reisizadeh2020fedpaq} (and related FedCOM~\cite{caldas2020fedcom}) combines periodic averaging with static top-\(k\) sparsification and quantization under theoretical guarantees, albeit at the cost of complex hyperparameter tuning. Recent \emph{dynamic sparsification} techniques adapt the communication support per round to gradient statistics: Zhang \emph{et al.}~\cite{zhang2022layer} proposed layer-wise mean thresholds in Adaptive Layerwise Threshold Sparsification, and Kumar \emph{et al.}~\cite{kumar2023noniid} introduced percentile-driven budgets in Non-IID Aware Sparse FL (ICML 2023). Concurrently, \emph{Error‐Feedback} has been refined for federated settings: Li \emph{et al.}~\cite{LiICLR2024} proved unbiased convergence under combined quantization, sparsification, and residual accumulation in Unified Convergence Analysis of Compressed FL, and Chen \emph{et al.}~\cite{ChenAAAI2025} introduce adaptive residual scaling in Stabilized Error-Feedback for FL. Our FedSparQ approach builds on these advances by unifying adaptive threshold sparsification, float16 quantization, and error-feedback into a hyperparameter-light framework that automatically balances communication reduction against convergence fidelity.

\textbf{Adaptive and Dynamic Compression.} Beyond fixed-$k$ or static bit-widths, adaptive compressors modulate rates according to signal statistics or resource constraints. \emph{AdaComp} adaptively selects residual components based on per-layer activity to tune compression on the fly \cite{chen2018adacomp}; \emph{SparseFed} applies sparsity-aware aggregation and robustness analysis in FL \cite{panda2022sparsefed}; and \emph{SoteriaFL} integrates stochastic compression with local differential privacy for federated settings \cite{li2022soteriafl}. FedSparQ follows this adaptive line via EMA-based layer-wise thresholds and error-feedback, while focusing on a lightweight design and half-precision coding of retained entries.

\section{FedSparQ framework}
\label{Sec:fedsparq}
\begin{figure*}[ht]
\scriptsize
\centering
\includegraphics[width=0.9\linewidth]{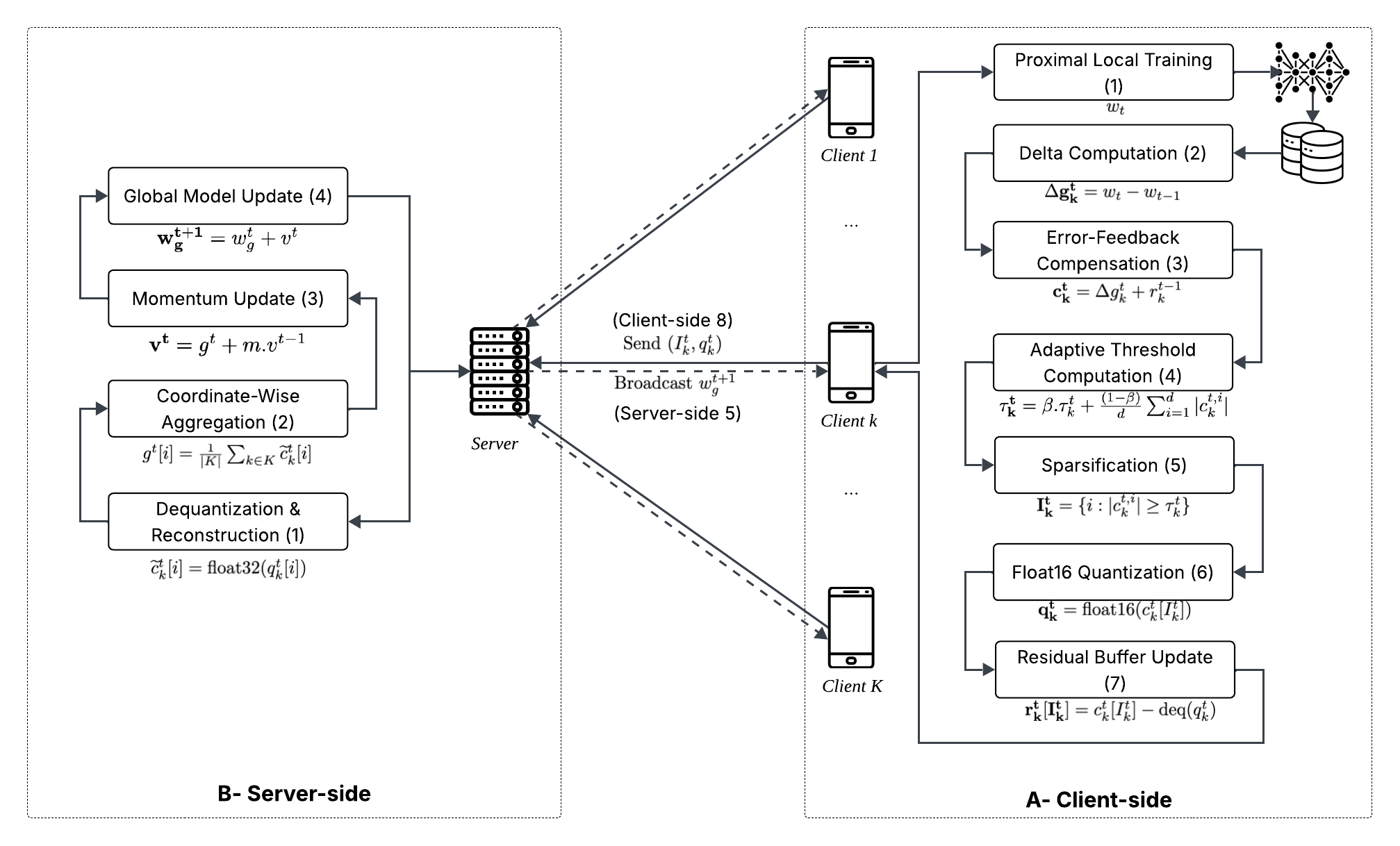}
\caption{FedSparQ system model.}
\label{fig:fedsparq}
\end{figure*}


FedSparQ is a two‐tier distributed framework comprising a central \emph{Parameter Server} and \(K\) edge \emph{Clients}. Figure~\ref{fig:fedsparq} illustrates the high‐level flow. FedSparQ introduces a lightweight, adaptive compression mechanism that removes the need for manual tuning of sparsity levels or quantization bit-widths. 
At each round, FedSparQ uses exponentially smoothed, per-layer thresholds (computed from error-feedback signals) to automatically select the most significant gradient coordinates. These selected values are then quantized to float16, significantly reducing communication costs while preserving important update information. Residuals are maintained in local buffers to ensure that no gradient information is permanently lost.
By integrating FedProx regularization, adaptive threshold-based sparsification, low-precision quantization, and server-side momentum into a unified pipeline, FedSparQ consistently achieves over 90\% communication reduction while maintaining near full-precision accuracy, even in highly non-IID and bandwidth-constrained settings.

\subsection{Client–Side Processing}

Each client \(k\) begins round \(t\) by receiving the global model parameters \(w^t_g\) from the server.  Internally, the client maintains:
\begin{itemize}
  \item A local model copy \(w^t\in\mathbb{R}^d\),  
  \item A residual buffer \(r^{\,t-1}_k\in\mathbb{R}^d\) that accumulates previously dropped updates.
\end{itemize}

\paragraph{Proximal Local Training (1) \& Delta Computation (2)}  
Upon setting \(w^t \leftarrow w^t_g\), the client trains for one or more local epochs using SGD (or an alternative optimizer) on its private data \(\mathcal{D}_k\).  After training, it computes the raw update  
\[
  \Delta g^t_k \;=\; w^t_k \;-\;w^{\,t-1}_k,
\]  
which captures the change in model parameters induced by the local data.

\paragraph{Error‐Feedback Compensation (3)}  
To correct for past sparsification bias, the client forms the compensated update:  
\[
  c^t_k \;=\;\Delta g^t_k + r^{\,t-1}_k.
\]  
This ensures that any coordinate previously omitted (and stored in \(r^{\,t-1}_k\)) is reintroduced into the current compression step.

\paragraph{Adaptive Threshold Sparsification (4 \& 5)}  
FedSparQ maintains a per-client threshold that evolves via exponential smoothing, following common practice in the sparse‐communication literature~\cite{lin2018deep}:  
\[
\bar s_k^t \;=\;\frac{1}{d}\sum_{i=1}^d\bigl|\mathbf c^{t,i}_k\bigr|,\qquad
\tau^t_k \;=\;\beta\,\tau^{t-1}_k \;+\;(1-\beta)\,\bar s_k^t,
\]  
where \(\beta\in[0,1]\) is a fixed smoothing constant  (\(\beta=0.9\)) imported from prior work on error‐feedback and gradient compression.  The support set is then  
\[
I_k^t = \bigl\{\,i : \lvert\mathbf c^{t,i}_k\rvert \ge \tau_k^t\bigr\}
\]  
automatically adapting to each client’s gradient magnitude while while smoothing out noise.

\paragraph{Float16 Quantization (6)}  
On the retained coordinates, the client applies half‐precision quantization:  
\[
  q^t_k = \mathrm{float16}\bigl(c^t_k[I^t_k]\bigr)
\]  
halving the per‐value byte cost from 4 to 2 bytes, while preserving enough precision for convergence.

\paragraph{Residual Buffer Update (7)}  
Finally, the client computes the new residual vector—zeros outside \(I^t_k\), and the dequantization error on \(I^t_k\):  
\[
 r^t_k[i] = 
 \begin{cases}
   c^t_{k,i} - \mathrm{float32}\bigl(q^t_{k,i}\bigr), & i\in I^t_k,\\
   0, & \text{otherwise}.
 \end{cases}
\]  
This residual will be added in the next round to ensure no information is permanently lost.

\paragraph{Sparse‐Quantized Payload Transmission (8)}  
Client \(k\) transmits the tuple \(\bigl(I^t_k, q^t_k\bigr)\) back to the server.  Since \(|I^t_k|\ll d\), and each index is 4 bytes while each quantized value is 2 bytes, the per‐round upload cost is:
\[
  \text{bytes} = 6\,|I^t_k| \;\ll\; 4d.
\]  
All the steps are detailed in algorithm~\ref{alg1:client}
\begin{algorithm}[h]
\caption{FedSparQ Client \(k\) (one round \(t\))}
\begin{algorithmic}[1]
  \State \textbf{Input:} Global model \(w^t_g\), past residual \(r^{t-1}_k\)
  \State \(w^t \gets w^t_g\)
  \State Train locally (e.g.\ one epoch of SGD) on \(\mathcal{D}_k\), yielding \(w^t_k\)
  \State \(\Delta g^t_k \gets w^t - w^{t-1}\) \hfill\emph{(local gradient computation)}
  \State \(c^t_k \gets \Delta g^t_k + r^{t-1}_k\) \hfill\emph{(error‐feedback)}
  \State \(\bar s_k^t \;=\;\frac{1}{d}\sum_{i=1}^d\bigl|\mathbf c^{t,i}_k\bigr|\)
  \State \(\tau^t_k \gets \beta\,\tau^{t-1}_k \;+\;(1-\beta)\,\bar s_k^t\)
  \State \(I^t_k \gets \{\,i:|c^t_{k,i}|\ge\tau^t_k\}\)
  \State \(q^t_k \gets \mathrm{float16}\bigl(c^t_k[I^t_k]\bigr)\)
  \State \(r^t_k \!\![I^t_k] \gets c^t_k[I^t_k] - \mathrm{float32}(q^t_k)\), else \(0\)
  \State \emph{return} Sparse payload \((I^t_k, q^t_k)\)
\end{algorithmic}
\label{alg1:client}
\end{algorithm}

\subsection{Server–Side Aggregation}

On the server, incoming sparse updates from the selected clients \(S_t\) are processed as follows:

\paragraph{Dequantization and Reconstruction (1)}  
For each client \(k\in S_t\), the server reconstructs a dense update vector  
\[
  \tilde c^t_k[i] = 
  \begin{cases}
    \mathrm{float32}\bigl(q^t_{k,i}\bigr), & i\in I^t_k,\\
    0, & \text{otherwise}.
  \end{cases}
\]  
This restores a full‐dimensional float32 update, with zeros filling non‐sent coordinates.

\paragraph{Coordinate‐Wise Averaging (2)}  
The server computes the global sparse gradient  
\[
  g^t[i] \;=\; \frac{1}{|S_t|}\sum_{k\in S_t} \tilde c^t_k[i],
  \quad i=1,\dots,d,
\]  
aggregating information across clients.  Coordinates that no client selected remain zero.

\paragraph{Momentum Update (3)}
To further stabilize and accelerate model global convergence, FedSparQ’s server augments its sparse‐gradient aggregation with a classical momentum term.

\[
\mathbf v^t \;=\;\mu\,\mathbf v^{t-1}\;+\;\mathbf g^t
\] with \[
\mathbf v_0 = \mathbf 0 \in \mathbb{R}^d.
\]
where \(\mu\in[0,1[\) (\(\mu=0.01\)) is the momentum constant.  This design borrows the same principle as momentum‐SGD.

\paragraph{Global Model Update (4)}  
The new global model parameters are obtained by a simple additive update:  
\[
  w^{t+1}_g = w^t_g + v^t
\]  
By using the reconstructed sparse update \(g^t\) rather than full‐precision deltas, the server achieves communication savings without sacrificing convergence speed.

\paragraph{Broadcast for Next Round (5)}  
Finally, \(w^{t+1}_g\) is broadcast to clients for the next round’s local processing, closing the training loop.

The server reconstructs client-sparse, quantized updates, averages only over current coordinates, applies momentum, and updates the global model. 

Algorithm~\ref{alg2:server} details all the steps the server does for the parameters aggregation

\begin{algorithm}[h]
\caption{FedSparQ Server (one round \(t\))}
\begin{algorithmic}[1]
  \State \textbf{Input:} Global model \(w^t_g\), $\mu = 0.01$, $v^{t-1}$ 
  \State Broadcast \(w^t_g\) to all clients \(k\in S_t\)
  \State Collect sparse payloads \(\{(I^t_k, q^t_k)\}\)
  \For{each client \(k\in S_t\)}
    \State \(\tilde c^t_{k}[i] \gets \begin{cases}\mathrm{float32}(q^t_{k,i}),&i\in I^t_k\\0,&\text{otherwise}\end{cases}\)
  \EndFor
  \State \(g^t \gets \tfrac1{|S_t|}\sum_{k\in S_t} \tilde c^t_k\)
    \State \(\mathbf v^t \gets \mu\,\mathbf v^{t-1} \;+\; \mathbf g^t\)
    \State \(\mathbf w^{t+1}_g \gets \mathbf w^t_g \;+\; \mathbf v^t\)
  \State \emph{return} \(w^{t+1}_g\)
\end{algorithmic}
\label{alg2:server}
\end{algorithm}

\subsection{Communication Analysis}
Each selected coordinate incurs 4 bytes (index) + 2 bytes (float16) = 6 bytes. Adaptive $|I_t|$ scales to gradient entropy, yielding average per-round bytes far below full-precision 32$d$.

\section{Convergence of FedSparQ}
\label{Sec:conv}
We analyze the round-$t$ update shown in Fig.~\ref{fig:fedsparq}. Each selected client $k\in K$ starts from the broadcast global model $w_g^{t}$, runs local FedProx steps to obtain parameters $w_k^{t}$, forms a compensated delta $\Delta g_k^{t}$, sparsifies it via an EMA threshold, quantizes the kept parameters to half precision, and transmits the pair $(I_k^t,\,q_k^t)$ to the server which dequantizes them to float32, averages, applies server momentum, and updates the global model.

\paragraph{Client-side update and compression.}
With FedProx coefficient $\mu\ge 0$, client $k$ performs $\tau$ proximal-SGD steps on
$f_k(w)+\tfrac{\mu}{2}\|w-w_g^{t}\|^2$ and computes the local model delta
\[
\Delta g_k^{t} = w_k^{t} - w_g^{t}.
\]
Error feedback adds the previous residual to obtain the compensated vector
\[
c_k^{t} = \Delta g_k^{t} + r_k^{t-1}.
\]
FedSparQ computes a \emph{layer-wise} EMA threshold; if a layer has $d$ coordinates, then
\[
\tau_k^{t} = \beta\,\tau_k^{t-1} + (1-\beta)\,\frac{1}{d}\sum_{i=1}^{d} |c_k^{t,i}|,\qquad \beta\in[0,1).
\]
We keep indices
\[
I_k^{t} = \{\, i : |c_k^{t,i}|\ge \tau_k^{t}\,\}.
\]
The kept values are quantized to half precision (with \emph{stochastic rounding} to ensure unbiasedness),
\[
q_k^{t} = \operatorname{float16}\!\big(c_k^{t}[I_k^{t}]\big),
\]
and the residual is updated by reinjecting the compression error using zero-padding at dropped indices:
\[
r_k^{t} = c_k^{t} - \tilde u_k^{t},\qquad
\tilde u_k^{t}[I_k^{t}] = \operatorname{deq}(q_k^{t}),\ \ \tilde u_k^{t}[\bar I_k^{t}] = 0.
\]

\paragraph{Server-side averaging and momentum.}
The server dequantizes each message $\tilde c_k^{t}=\operatorname{float32}(q_k^{t})$ and averages
\[
g^{t}[i] = \frac{1}{|K|}\sum_{k\in K} \tilde c_k^{t}[i].
\]
With momentum $m\in[0,1)$ it computes $v^{t} = g^{t} + m\,v^{t-1}$ and updates
\[
w_g^{t+1} = w_g^{t} + v^{t}.
\]

\paragraph{Assumptions (noise model)}
\begin{itemize}
    \item (A1) Each $f_k$ is $L$-smooth ($L$ is the smoothness constant a.k.a. the Lipschitz constant).
    \item (A2) Stochastic gradients are unbiased with variance $\le \sigma^2$.
    \item (A3) The threshold sparsifier is \emph{contractive}:   
    $\mathbb{E}\|x-\mathcal{C}(x)\|^2 \le (1-\delta)\|x\|^2$ for some $\delta\in]0,1]$ (here $\delta$ equals the layer keep fraction; EMA thresholds guarantee a per-round lower bound $\delta\ge\delta_{\min}>0$).
    \item (A4) The float16 coder with stochastic rounding is an \emph{unbiased} quantizer with variance parameter $\beta\ge 0$: $\mathbb{E}[\mathcal{Q}(x)]=x$ and $\mathbb{E}\|\mathcal{Q}(x)-x\|^2 \le \beta\|x\|^2$.
    \item (A5) FedProx’s bounded-dissimilarity condition holds for non-IID clients.
\end{itemize}

\paragraph{Lemma (EF telescoping under FedSparQ)}
Under (A3),
\[
\mathbb{E}\|r_k^{t}\|^2 \le (1-\delta)\,\mathbb{E}\|r_k^{t-1}\|^2 + (1-\delta)\,\mathbb{E}\|\Delta g_k^{t}\|^2,
\]
\[
\quad\text{and}\quad
c_k^{t} - \tilde u_k^{t} = r_k^{t}-r_k^{t-1}
\]
Hence sparsification bias cancels in the \emph{cumulative} update, only variance-like terms remain.

\paragraph{Theorem (Non-convex rate)}
Supposing the assumptions (A1)–(A5) hold and choose a constant step size $\eta\le 1/(2L)$
(allowing a slightly smaller constant when momentum is used).
Let $G^2 \ge \sup_t \mathbb{E}\|\Delta g_k^{t}\|^2$.
Then

\begin{multline*}
\frac{1}{T}\sum_{t=0}^{T-1}\mathbb{E}\|\nabla f(w_g^{t})\|^2
\;\le\;
\frac{2\big(f(w_g^{0})-f^*\big)}{\eta T} 
\;+\;
\frac{3L\eta}{|K|}\,\sigma^2 \\
\;+\;
6L\eta\,\Big(\tfrac{1-\delta_{\min}}{\delta_{\min}}\Big)\,G^2
\;+\;
6L\eta\,\beta\,G^2
\end{multline*}

With $\eta=\Theta(1/\sqrt{T})$ which is the standard $\tilde{\mathcal{O}}(1/\sqrt{T})$ decay; the last two terms quantify the price of sparsity (via $\delta_{\min}$) and quantization (via $\beta$).

\paragraph{Corollary (Polyak–Łojasiewicz~\cite{karimi2016})}
If $f$ satisfies the Polyak–Łojasiewicz condition with constant $\mu_{\text{PL}}>0$,
then for a suitable constant $\eta$ the iterates converge linearly to a neighborhood
of size $\mathcal{O}\!\big((\tfrac{1-\delta_{\min}}{\delta_{\min}}+\beta)G^2 + \sigma^2|K|\big)$.

\paragraph{\emph{FedSparQ} convergence}

(i) EMA thresholds keep the effective keep-rate bounded away from $0$ (i.e., $\delta_{\min}>0$),
\\
(ii) using stochastic rounding for float16 satisfies (A4); deterministic rounding can be treated as bounded bias adding a small constant term,
\\
(iii) server momentum mainly affects constants and allowable step sizes,
\\
(iv) the analysis is per-layer and uses the worst-case keep fraction across layers.

Under standard smoothness and bounded-noise assumptions mentioned previously, error feedback residual ($r_k^t$) removes sparsification bias, so FedSparQ attains the usual non-convex $\mathcal{O}(1/\sqrt{T})$ rate; under a Polyak–Łojasiewicz condition it converges linearly to a small neighborhood determined by sparsity and quantization noise.




\section{Results and discussion}
\label{Sec:Results}
This section presents an empirical evaluation of FedSparQ across different baselines and three datasets under IID and non-IID conditions, focusing on the accuracy–bandwidth–stability trade-off via communication cost, performance, convergence and robustness.
FedSparQ shows substantial communication reductions while maintaining similar or better baseline accuracy.

\begin{figure}[t]
\centering
\pgfplotsset{
every axis/.append style={
font=\scriptsize,
line width=1pt},
scale only axis,
}
\usetikzlibrary{patterns}
\begin{tikzpicture}
\begin{axis}[
    ybar,
    bar width=0.2cm,
    ylabel={Total Bytes sent},
    symbolic x coords={MNIST, FMNIST, CIFAR},
    xtick=data,
    x tick label style={anchor=north},
    legend style={at={(0.5,1.05)}, anchor=south, legend columns=3, font=\scriptsize},
    width=7cm,
    height=6cm,
ymode=log,
    ymajorgrids=true,
legend entries={FedAvg, FedPAQ, FedSparQ, Quantization, Random sparse $10\%$, Static top $10\%$ sparse}, 
    enlarge x limits=0.25,
]
\addplot [fill=green] coordinates {(MNIST, 10177000) (FMNIST,185013000) (CIFAR,4431747672)};
\addplot [fill=purple] coordinates {(MNIST, 300000) (FMNIST,72000000) (CIFAR,304000000)};
\addplot [fill=red] coordinates {(MNIST, 6240386.66666667) (FMNIST,34385013.3333333) (CIFAR,472171610)};
\addplot [fill=cyan] coordinates {(MNIST, 5088500) (FMNIST,92506300) (CIFAR,2260635004)};
\addplot [fill=orange] coordinates {(MNIST,300000) (FMNIST,72000000) (CIFAR,304000000)};
\addplot [fill=teal] coordinates {(MNIST,300000) (FMNIST,72000000) (CIFAR,323342524)};

\end{axis}
\label{fig:communicationcost}
\end{tikzpicture}
\caption{Communication Cost (Bytes sent) on CIFAR-10, FMNIST and MNIST.}
\label{fig:cifar_bytes_iid}
\end{figure}
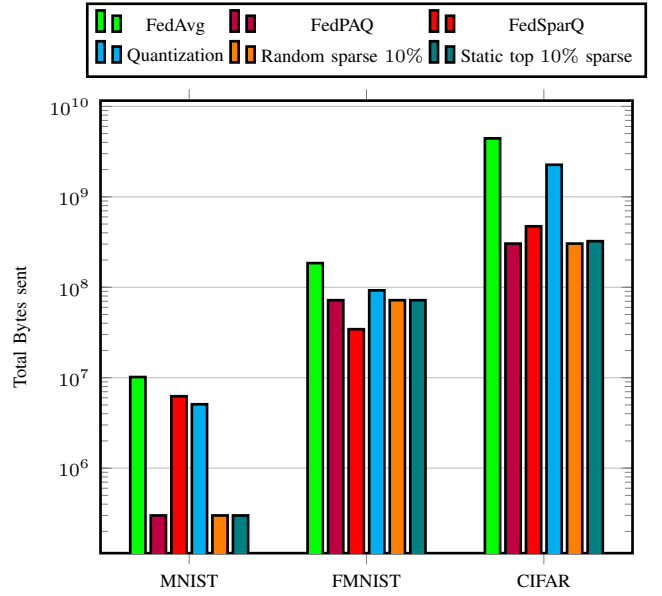

\begin{figure*}[t!]
\centering
\begin{subfigure}[t]{0.3\linewidth}
\pgfplotsset{
every axis/.append style={
font=\scriptsize,
line width=1pt},
scale only axis,
}
\pgfplotsset{
compat=1.11,
legend image code/.code={
\draw[mark repeat=2,mark phase=2]
plot coordinates {
(0cm,0cm)
(0.1cm,0cm)        
(0.3cm,0cm)         
};%
}
}
\begin{tikzpicture}
\begin{axis}[
ylabel={Accuracy},
xlabel={Step},
grid=major,
xmin=0,
xmax=105,
width=4.5cm,
height=3cm,
legend entries={FedAvg, FedPAQ, FedSparQ, Quantization, Random sparse $10\%$, Static top $10\%$ sparse}, 
 legend columns = 2,
 legend style = {at={((0.5,1.05)}, anchor=south, font=\tiny},
]
\addplot [green] table [y index=1, x index=0] {results/values_converted/fedavg_acc_cifar_iid.csv};
\addplot [purple] table [y index=1, x index=0] {results/values_converted/fedpaq_acc_cifar_iid.csv};

\addplot [red] table [y index=1, x index=0] {results/values_converted/hybrid_acc_cifar_iid.csv};
\addplot [cyan] table [y index=1, x index=0] {results/values_converted/quant_acc_cifar_iid.csv};
\addplot [orange] table [y index=1, x index=0] {results/values_converted/ran_acc_cifar_iid.csv};
\addplot [teal] table [y index=1, x index=0] {results/values_converted/static_acc_cifar_iid.csv};

\end{axis}

\end{tikzpicture}
\caption{CIFAR-10 (IID)}\label{subfig:accuracy_cifar10}
\end{subfigure}
\hfill
\begin{subfigure}[t]{0.3\linewidth}
\pgfplotsset{
every axis/.append style={
font=\scriptsize,
line width=1pt},
scale only axis
}
\pgfplotsset{
compat=1.11,
legend image code/.code={
\draw[mark repeat=2,mark phase=2]
plot coordinates {
(0cm,0cm)
(0.1cm,0cm)        
(0.3cm,0cm)         
};%
}
}

\begin{tikzpicture}
\begin{axis}[
ylabel={Accuracy},
xlabel={Step},
grid=major,
xmin=0,
xmax=26,
width=4.5cm,
height=3cm,
legend entries={FedAvg, FedPAQ, FedSparQ, Quantization, Random sparse $10\%$, Static top $10\%$ sparse}, 
 legend columns = 2,
 legend style = {at={((0.5,1.05)}, anchor=south, font=\tiny, mark options={scale=0.5}
},
]
\addplot [green] table [y index=1, x index=0] {results/values_converted/fedavg_acc_fmnist_iid.csv};
\addplot [purple] table [y index=1, x index=0] {results/values_converted/fedpaq_acc_fmnist_iid.csv};
\addplot [red] table [y index=1, x index=0] {results/values_converted/hybrid_acc_fmnist_iid.csv};
\addplot [cyan] table [y index=1, x index=0] {results/values_converted/quant_acc_fmnist_iid.csv};
\addplot [orange] table [y index=1, x index=0] {results/values_converted/ran_acc_fmnist_iid.csv};
\addplot [teal] table [y index=1, x index=0] {results/values_converted/static_acc_fmnist_iid.csv};

\end{axis}

\end{tikzpicture}
\caption{Fashion-MNIST (IID)}\label{subfig:accuracy_fmnist}
\end{subfigure}
\hfill
\begin{subfigure}[t]{0.35\linewidth}
\pgfplotsset{
every axis/.append style={
font=\scriptsize,
line width=1pt},
scale only axis,
}
\pgfplotsset{
compat=1.11,
legend image code/.code={
\draw[mark repeat=2,mark phase=2]
plot coordinates {
(0cm,0cm)
(0.1cm,0cm)        
(0.3cm,0cm)         
};%
}
}
\begin{tikzpicture}
\begin{axis}[
ylabel={Accuracy},
xlabel={Step},
grid=major,
xmin=0,
xmax=26,
width=4.5cm,
height=3cm,
legend entries={FedAvg, FedPAQ, FedSparQ, Quantization, Random sparse $10\%$, Static top $10\%$ sparse}, 
 legend columns = 2,
 legend style = {at={((0.5,1.05)}, anchor=south, font=\tiny},
]
\addplot [green] table [y index=1, x index=0] {results/values_converted/fedavg_acc_mnist.csv};
\addplot [purple] table [y index=1, x index=0] {results/values_converted/fedpaq_acc_mnist.csv};

\addplot [red] table [y index=1, x index=0] {results/values_converted/hybrid_acc_mnist.csv};
\addplot [cyan] table [y index=1, x index=0] {results/values_converted/quant_acc_mnist.csv};
\addplot [orange] table [y index=1, x index=0] {results/values_converted/ran_acc_mnist.csv};
\addplot [teal] table [y index=1, x index=0] {results/values_converted/static_acc_mnist.csv};

\end{axis}

\end{tikzpicture}
\caption{MNIST}\label{subfig:accuracy_mnist}
\end{subfigure}
\caption{Test accuracy vs. rounds on different dataset.}
\label{fig:Accuracy}
\end{figure*}
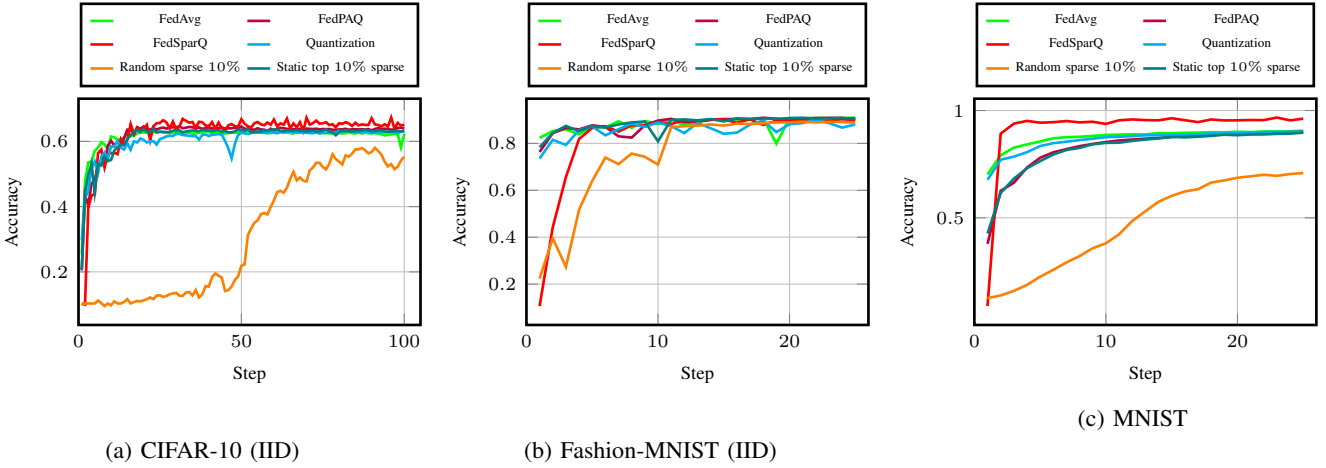

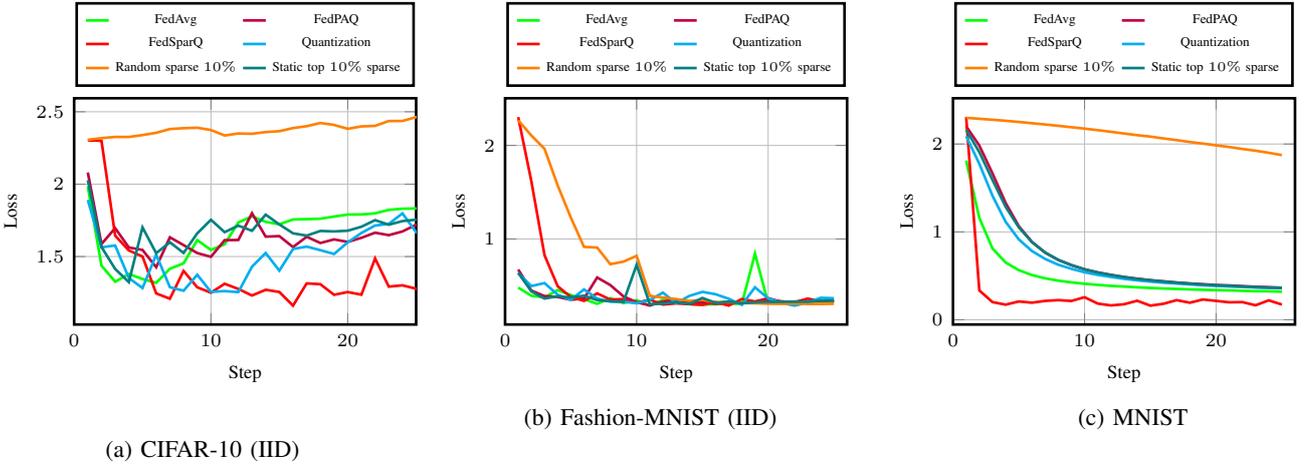
\begin{figure*}[t!]
\centering
\begin{subfigure}[t]{0.3\linewidth}
\pgfplotsset{
every axis/.append style={
font=\scriptsize,
line width=1pt},
scale only axis,
}
\pgfplotsset{
compat=1.11,
legend image code/.code={
\draw[mark repeat=2,mark phase=2]
plot coordinates {
(0cm,0cm)
(0.1cm,0cm)        
(0.3cm,0cm)         
};%
}
}
\begin{tikzpicture}
\begin{axis}[
ylabel={Loss},
xlabel={Step},
grid=major,
xmin=0,
xmax=25,
width=4.5cm,
height=3cm,
legend entries={FedAvg, FedPAQ, FedSparQ, Quantization, Random sparse $10\%$, Static top $10\%$ sparse}, 
 legend columns = 2,
 legend style = {at={((0.5,1.05)}, anchor=south, font=\tiny},
]
\addplot [green] table [y index=1, x index=0] {results/values_converted/fedavg_loss_cifar_iid.csv};
\addplot [purple] table [y index=1, x index=0] {results/values_converted/fedpaq_loss_cifar_iid.csv};

\addplot [red] table [y index=1, x index=0] {results/values_converted/hybrid_loss_cifar_iid.csv};
\addplot [cyan] table [y index=1, x index=0] {results/values_converted/quant_loss_cifar_iid.csv};
\addplot [orange] table [y index=1, x index=0] {results/values_converted/ran_loss_cifar_iid.csv};
\addplot [teal] table [y index=1, x index=0] {results/values_converted/static_loss_cifar_iid.csv};

\end{axis}

\end{tikzpicture}
\caption{CIFAR-10 (IID)}\label{subfig:loss_cifar10}
\end{subfigure}
\hfill
\begin{subfigure}[t]{0.3\linewidth}
\pgfplotsset{
every axis/.append style={
font=\scriptsize,
line width=1pt},
scale only axis,
}
\pgfplotsset{
compat=1.11,
legend image code/.code={
\draw[mark repeat=2,mark phase=2]
plot coordinates {
(0cm,0cm)
(0.1cm,0cm)        
(0.3cm,0cm)         
};%
}
}
\begin{tikzpicture}
\begin{axis}[
ylabel={Loss},
xlabel={Step},
grid=major,
xmin=0,
xmax=26,
width=4.5cm,
height=3cm,
legend entries={FedAvg, FedPAQ, FedSparQ, Quantization, Random sparse $10\%$, Static top $10\%$ sparse}, 
 legend columns = 2,
 legend style = {at={((0.5,1.05)}, anchor=south, font=\tiny},
]
\addplot [green] table [y index=1, x index=0] {results/values_converted/fedavg_loss_fmnist_iid.csv};
\addplot [purple] table [y index=1, x index=0] {results/values_converted/fedpaq_loss_fmnist_iid.csv};

\addplot [red] table [y index=1, x index=0] {results/values_converted/hybrid_loss_fmnist_iid.csv};
\addplot [cyan] table [y index=1, x index=0] {results/values_converted/quant_loss_fmnist_iid.csv};
\addplot [orange] table [y index=1, x index=0] {results/values_converted/ran_loss_fmnist_iid.csv};
\addplot [teal] table [y index=1, x index=0] {results/values_converted/static_loss_fmnist_iid.csv};

\end{axis}

\end{tikzpicture}
\caption{Fashion-MNIST (IID)}\label{subfig:loss_fmnist}
\end{subfigure}
\hfill
\begin{subfigure}[t]{0.35\linewidth}
\pgfplotsset{
every axis/.append style={
font=\scriptsize,
line width=1pt},
scale only axis,
}
\pgfplotsset{
compat=1.11,
legend image code/.code={
\draw[mark repeat=2,mark phase=2]
plot coordinates {
(0cm,0cm)
(0.1cm,0cm)        
(0.3cm,0cm)         
};%
}
}

\begin{tikzpicture}
\begin{axis}[
ylabel={Loss},
xlabel={Step},
grid=major,
xmin=0,
xmax=26,
width=4.5cm,
height=3cm,
legend entries={FedAvg, FedPAQ, FedSparQ, Quantization, Random sparse $10\%$, Static top $10\%$ sparse}, 
 legend columns = 2,
 legend style = {at={((0.5,1.05)}, anchor=south, font=\tiny},
]
\addplot [green] table [y index=1, x index=0] {results/values_converted/fedavg_loss_mnist.csv};
\addplot [purple] table [y index=1, x index=0] {results/values_converted/fedpaq_loss_mnist.csv};

\addplot [red] table [y index=1, x index=0] {results/values_converted/hybrid_loss_mnist.csv};
\addplot [cyan] table [y index=1, x index=0] {results/values_converted/quant_loss_mnist.csv};
\addplot [orange] table [y index=1, x index=0] {results/values_converted/ran_loss_mnist.csv};
\addplot [teal] table [y index=1, x index=0] {results/values_converted/static_loss_mnist.csv};

\end{axis}

\end{tikzpicture}
\caption{MNIST}\label{subfig:loss_mnist}
\end{subfigure}
\caption{Test loss vs. rounds on different dataset.}
\label{fig:Loss}
\end{figure*}
\subsection{Experimental Settings}
We have conducted a comprehensive experiment to validate the FedSparQ in terms of robustness and performance under different conditions. We have compared FedSparQ against five state-of-the-art baseline methods: full-precision FedAvg, FedPAQ, standard quantisation, static Top-$10\%$ sparsification, and random Top-$10\%$ sparsification. Our experiments were conducted over three widely adopted benchmark datasets with varying complexity levels and architectural requirements, including: (i) CIFAR-10 with ResNet-18 architecture trained over 100 communication rounds across 3 clients, (ii) Fashion-MNIST with LeNet-5 architecture trained over 25 rounds across 3 clients, and (iii) MNIST with a multi-layer perceptron (MLP) trained over 25 rounds across 3 clients. We evaluate the model's robustness under both IID and non-IID distributions. Uniform random sampling was used for IID experiments to ensure each client received an equal proportion of the dataset with balanced class representation. While for non-IID experiments, we simulate realistic federated environments by partitioning data using a Dirichlet distribution with concentration parameter ($\alpha=0.5$), which induces significant statistical heterogeneity across clients while maintaining sufficient data diversity for convergence.
Finally, we evaluate algorithm performance across four critical dimensions: (i) per-round communication overhead measured in bytes transmitted, (ii) test accuracy on held-out datasets, (iii) test loss convergence characteristics, and (iv) two complementary robustness metrics (see Equation \ref{eq:robust1} and \ref{eq:robust2}) that capture system resilience under data heterogeneity:

\begin{align}
\mathrm{Robust}_1 &= \mathrm{Accuracy} - \mathrm{Loss}, \label{eq:robust1}\\[4pt]
\mathrm{Robust}_2 &= \frac{\mathrm{Accuracy}}{\mathrm{Loss} + \varepsilon}, \qquad \varepsilon > 0 \label{eq:robust2}
\end{align}

(with $\epsilon \approx 0$ to prevent division by zero).  
These robustness scores capture the trade-off between predictive performance and stability under distributional shift.

\subsection{FedSparQ performance and communication–efficiency under IID Trade-off}
FedSparQ consistently delivers state-of-the-art accuracy at a fraction of the communication cost under IID data distribution, with smooth and fast convergence. Table~\ref{tab:iid_results} and Figures~\ref{fig:cifar_bytes_iid}–\ref{fig:Loss} report the communication cost (bytes uploaded), final test accuracy, and final loss for all methods on MNIST, Fashion-MNIST, and CIFAR-10 under IID partitions.

\begin{table*}[t]
\centering
\scriptsize
\caption{IID results on MNIST, FMNIST and CIFAR-10: total upload bytes, final test accuracy (\%) and final training loss.}
\label{tab:iid_results}
\small
\begin{adjustbox}{width=\textwidth,center}
\begin{tabular}{l 
   >{\hfil}c<{\hfil} >{\hfil}c<{\hfil} >{\hfil}c<{\hfil} 
   >{\hfil}c<{\hfil} >{\hfil}c<{\hfil} >{\hfil}c<{\hfil} 
   >{\hfil}c<{\hfil} >{\hfil}c<{\hfil} >{\hfil}c<{\hfil}}
\toprule
 & \multicolumn{3}{c}{\textbf{MNIST}} 
 & \multicolumn{3}{c}{\textbf{FMNIST}} 
 & \multicolumn{3}{c}{\textbf{CIFAR-10}} \\
\cmidrule(lr){2-4} \cmidrule(lr){5-7} \cmidrule(lr){8-10}
\textbf{Method}
 & \textbf{Bytes}
 & \textbf{Acc}
 & \textbf{Loss}
 & \textbf{Bytes}
 & \textbf{Acc}
 & \textbf{Loss}
 & \textbf{Bytes}
 & \textbf{Acc}
 & \textbf{Loss} \\
\midrule
FedAvg
 & $1.02\times10^7$ & $90.5$ & $0.33$
 & $1.85\times10^8$ & $\mathbf{91.8}$ & $0.31$
 & $4.69\times10^9$ & $62.25$ & $1.47$ \\

FedPAQ
 & $\mathbf{3.04\times10^5}$ & $91.0$ & $0.36$
 & $7.25\times10^7$ & $89.5$ & $0.37$
 & $\mathbf{3.04\times10^8}$ & $63.03$ & $1.94$ \\

\textbf{FedSparQ}
 & $6.24\times10^5$ & $\mathbf{96.3}$ & $\mathbf{0.17}$
 & $\mathbf{3.35\times10^7}$ & $91.2$ & $0.33$
 & $4.69\times10^8$ & $\mathbf{64.85}$ & $\mathbf{1.41}$\\

Quantization
 & $5.08\times10^6$ & $89.2$ & $0.37$
 & $9.25\times10^7$ & $89.4$ & $0.37$
 & $2.25\times10^9$ & $63.03$ & $1.90$ \\

Random 10\% Sparse
 & $\mathbf{3.04\times10^5}$ & $71.8$ & $1.88$
 & $7.20\times10^7$ & $89.8$ & $\mathbf{0.31}$
 & $3.00\times10^8$ & $55.23$ & $1.45$ \\

Static Top 10\%
 & $\mathbf{3.04\times10^5}$& $89.6$ & $0.36$
 & $7.20\times10^7$ & $90.54$ & $0.36$
 & $3.15\times10^8$ & $62.9$ & $2.05$ \\
\bottomrule
\end{tabular}
\end{adjustbox}
\end{table*}

\begin{table*}[t!]
\centering
\scriptsize
\caption{Non-IID Results on FMNIST and CIFAR-10: Accuracy, Loss, and Robustness Metrics.}
\label{tab:niid_both}
\begin{adjustbox}{width=\textwidth,center}
\begin{tabular}{l
    S[table-format=2.1] S[table-format=1.2] S[table-format=2.2] S[table-format=2.2]
    S[table-format=2.1] S[table-format=1.2] S[table-format=2.2] S[table-format=2.2]
}
\toprule
 & \multicolumn{4}{c}{FMNIST (non-IID)} 
 & \multicolumn{4}{c}{CIFAR-10 (non-IID)} \\
\cmidrule(lr){2-5} \cmidrule(lr){6-9}
Method
  & {Acc (\%)} & {Loss} & {Robust$_1$} & {Robust$_2$}
  & {Acc (\%)} & {Loss} & {Robust$_1$} & {Robust$_2$}
\\
\midrule
FedAvg            
  & 87.26   & 0.39   & 86.87   & 223.74  
  & 54.46   & 3.09   & 51.37   & 17.62  
\\
FedPAQ            
  & 86.23   & 0.42   & 85.81   & 205,31  
  & 31.05  & 2.23   & 28.82   & 13.92  
\\
\bfseries FedSparQ 
  & 87.26  & 0.48 & 86.78   & 181.79  
  & \bfseries{66.3}   & \bfseries{1.51}   & \bfseries{64.79}   & \bfseries{43.91} 
\\
Quantization      
  & \bfseries{90.64}   & \bfseries{0.29}   & \bfseries{90.64}   & \bfseries{312.55}  
  & 54.65   & 2.43   & 52.22   & 22.48  
\\
Random 10\% Sparse
  & 89.33   & 0.30  & 89.03   & 297.76  
  & 44.4   & 1.66   & 42.74   & 26.74  
\\
Static Top 10\%   
  & 85.56   & 0.49   & 85.07   & 174.61 
  & 54.8   & 2.31   & 52.49   & 23.72  
\\
\bottomrule
\end{tabular}%
\end{adjustbox}
\end{table*}

As reported in Table~\ref{tab:iid_results}, \textbf{FedSparQ} outperform other baseline methods with accuracy/loss (96.3\% / 0.17) while transmitting only $6.24\times 10^{5}$ bytes (see Table~\ref{tab:iid_results} and Figure \ref{fig:cifar_bytes_iid}), over $16\times$ less than the full-precision FedAvg configuration ($1.02\times 10^{7}$) using the MNIST dataset. It reaches $\sim 80\%$ accuracy by round 3 and saturates by round 8 (see Figure \ref{subfig:accuracy_mnist}), evidencing low optimization bias from sparsification. Random Top-10\% collapses to 71.8\% (loss 1.88), underscoring the need for data-aware selection rather than unstructured pruning. FedPAQ matches the lowest byte budget but trails markedly in accuracy. We have noticed FedSparQ behaves differently over the FMNIST dataset, where it transmits $\approx 3.35\times 10^7$ bytes, which is roughly $3\times$ less than quantization and $5\times$ less than FedAvg, while obtaining good accuracy with a slight gap around 0.6\% compared to FedAvg’s 91.8\%, and maintaining a final loss of 0.33. Learning curves remain well-behaved and closely track the dense baseline (see Figure \ref{subfig:accuracy_fmnist}), showing that adaptive sparsity preserves convergence on moderate complexity vision tasks at a fraction of the bandwidth. 
Moreover, FedSparQ was tested over the CIFAR-10 dataset, obtaining the highest accuracy (64.85\%) and the lowest loss (1.41) among all methods (see Figures \ref{subfig:accuracy_cifar10} , \ref{subfig:loss_cifar10} and Table~\ref{tab:iid_results}), and transmits $4.69\times 10^8$ bytes. FedPAQ reduces traffic but converges to higher losses, 1.94, and lower accuracies. At the same time, random/static sparsity degrades further, evidence that \emph{adaptive} selection plus error-feedback is required when noisy gradients are high-variance. These results are notable because CIFAR10 is particularly sensitive to update bias and information loss, which are mitigated by FedSparQ’s adaptive threshold and residual feedback mechanisms.

\begin{figure*}[ht]
\centering
\includegraphics[width=0.4\linewidth]{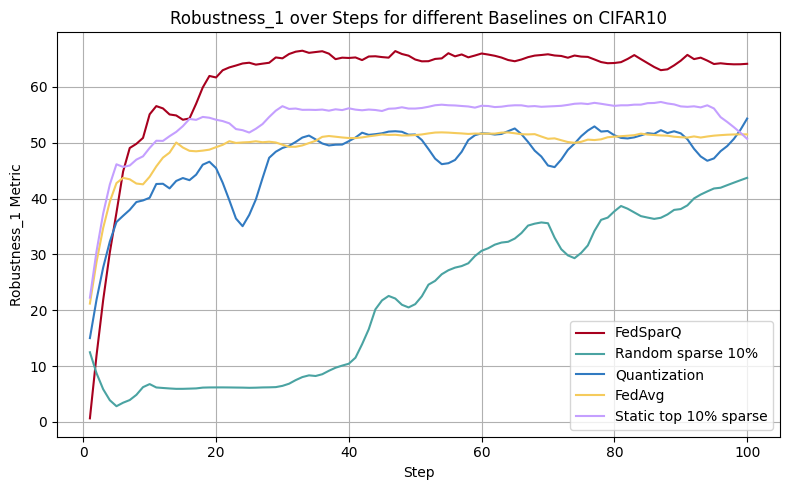}
\includegraphics[width=0.4\linewidth]{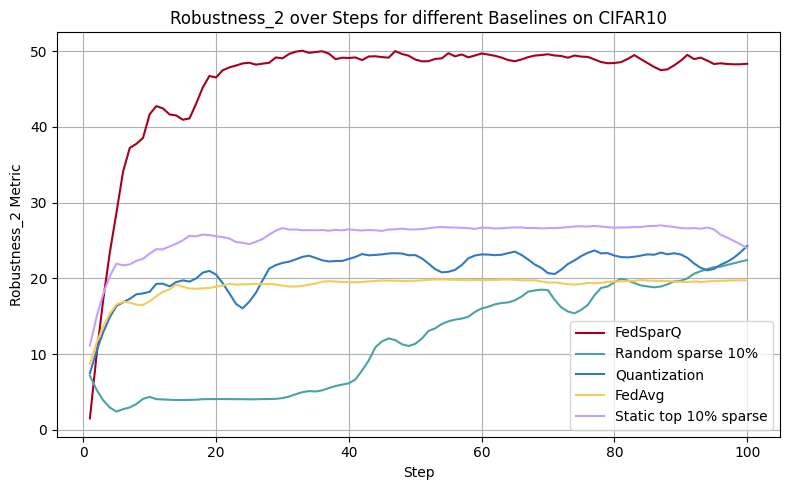}
\includegraphics[width=0.4\linewidth]{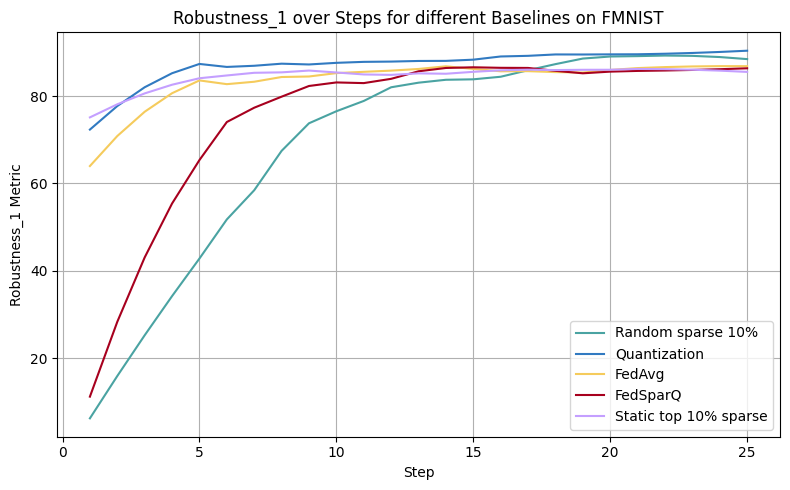}
\includegraphics[width=0.4\linewidth]{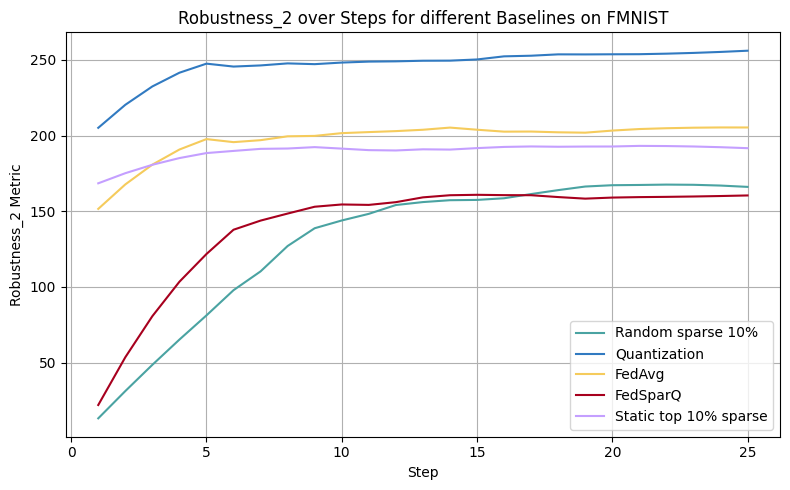}
\caption{Robustness under Non-IID data on CIFAR-10 and FMNIST datasets.}
\label{fig:robust}
\end{figure*}


\subsection{Robustness and Stability under Statistical Heterogeneity}
\label{sec:robustness}

We further evaluate all methods' robustness and stability over FMNIST and CIFAR10 datasets using Dirichlet ($\alpha=0.5$) non-IID splits.

 As shown in Table~\ref{tab:niid_both}, Quantization achieves the highest absolute accuracy (90.64\%) and robustness scores (90.64\%) over the FMNIST dataset. FedSparQ and FedAvg remain in a good position with a slight gap in raw accuracy (87.26\%). While random sparsity ranked in the second place, as it's clearly observed in both robustness metrics, despite sending no extra metadata and using a strict byte budget (see Table~\ref{tab:niid_both} and Figure~\ref{fig:robust}). 
 
 Moreover, evaluating all methods over the CIFAR-10 benchmark, a more complex and diverse dataset, FedSparQ achieves the best overall performance among all evaluated methods (see Figure~\ref{fig:robust} and Table~\ref{tab:niid_both}). Specifically, it achieves 66.3\% top-1 accuracy with the lowest loss (1.51) and the highest robustness scores ($ \text{Robust}_1\approx64.8$, $ \text{Robust}_2\approx43.9$). In this harsher scenario, where client drift and gradient variance are visible, FedSparQ’s combination of adaptive thresholding and residual error-feedback effectively reduces information loss across rounds and prevents the destabilization observed with static or purely quantized schemes. 

Based on these results, we observed the following behaviour of \textbf{FedSparQ}:  
\begin{itemize}
    \item Matches or exceeds the accuracy of FedAvg while using \textbf{23\% less bandwidth} than float16 quantization and up to \textbf{90\% less} than full-precision.
    \item Consistently outperforms static sparsification and random dropping, both in IID and non-IID regimes.
    \item Achieves strong robustness in non-IID settings, particularly on CIFAR-10, where adaptive sparsification is crucial.
    \item Requires only a simple threshold rule as a hyperparameter, avoiding the delicate tuning demanded by static methods.
\end{itemize}

The obtained results indicated that adaptive communication combined with residual (error) feedback provides a principled and practical foundation for FL communication efficiency. By adjusting sparsification/quantization thresholds to the evolving signal and recycling untransmitted gradient information across rounds, this design preserves informative updates, reduces client-drift–induced instability, and maintains convergence under apparent data heterogeneity. These mechanisms explain FedSparQ’s consistent gains over all baselines, simultaneously improving accuracy and lowering loss while enhancing robustness, thereby highlighting adaptivity at the communication/compression layer as a key lever for stable, high-performance FL in non-IID settings.

The results and discussion of the paper highlight that FedSparQ consistently achieves substantial communication savings while maintaining or even improving model accuracy across multiple datasets and settings. Compared to full-precision FedAvg, FedSparQ reduces bandwidth usage by up to 90\% and converges faster with smoother accuracy and loss curves. Under IID conditions, it matches or outperforms all baselines, while under non-IID distributions, especially on the challenging CIFAR-10 dataset, it shows superior robustness and stability due to its adaptive thresholding and error-feedback mechanisms. 

Unlike static sparsification or naive quantization, FedSparQ effectively balances efficiency and reliability without heavy hyperparameter tuning, proving its practicality for heterogeneous federated deployments. This trend suggests that adaptive, feedback-driven compression frameworks will be central to future federated learning systems, potentially combined with privacy-enhancing methods such as secure aggregation and differential privacy to achieve both scalability and trustworthiness. A future direction would be to investigate variational inference \cite{FL1}, especially under data drift \cite{FL2, FL3}.

\section{Conclusion}
\label{Sec:conclu}
In this paper, we proposed FedSparQ, a hyperparameter‐light compression framework for federated learning that combines adaptive per‐round threshold sparsification, float16 quantization, and error‐feedback residuals to eliminate communication bottlenecks without sacrificing the model's performance. Through extensive experiments on MNIST, Fashion-MNIST, and CIFAR-10 under both IID and non-IID partitions of data, we showed that FedSparQ cuts upload traffic by up to $10\times$ versus FedAvg while matching or exceeding its final test performance and substantially improving robustness across clients. In future works, we will extend FedSparQ with client-adaptive precision scheduling, provide theoretical convergence guarantees under combined compression and non-convex optimization, and explore its integration with privacy-preserving protocols such as secure aggregation and differential privacy.

\section\*{Acknowledgment}
This work was funded in whole or in part by the Luxembourg National Research Fund (FNR) LightGridSEED Project, ref. C21/IS/16215802/LightGridSEED .

\balance

\bibliographystyle{IEEEtran}

\end{document}